%% file: neurips_2026.tex
\documentclass{article}

 \usepackage[preprint]{neurips_2026}

\usepackage[utf8]{inputenc} 
\usepackage[T1]{fontenc}    
\usepackage{hyperref}       
\usepackage{url}            
\usepackage{booktabs}       
\usepackage{amsfonts}       
\usepackage{nicefrac}     
\usepackage{graphicx}
\usepackage{microtype}      
\usepackage{xcolor}         
\usepackage{pifont}
\usepackage{multirow}
\usepackage{makecell}
\usepackage{subcaption}
\usepackage{makecell} 
\usepackage{amsmath}
\usepackage{enumitem}
\title{AnyMo: Scaling Any-Modality Conditional Motion Generation with Masked Modeling}

\author{%
  \small \textbf{Yiheng Li}$^{1,2}$, \textbf{Zhuo Li}$^{3}$\thanks{Project leader}, \textbf{Ruibing Hou}$^{1}$\thanks{Corresponding author} \;, \textbf{Yingjie Chen}$^{3}$, \textbf{Hong Chang}$^{1,2}$, \textbf{Hao Liu}$^{3}$, \textbf{Shiguang Shan}$^{1,2}$ \\ 
  \small {$^1$Key Laboratory of Intelligent Information Processing of Chinese Academy of Sciences (CAS),} \\ \small {Institute of Computing Technology, CAS, China} \\
  \small {$^2$University of Chinese Academy of Sciences, China} \\
  \small {$^3$Independent Author} \\ 
  \small \texttt{\{yiheng.li,zhuo.li\}@vipl.ict.ac.cn,\{houruibing,changhong,sgshan\}@ict.ac.cn}\\
  \small \texttt{chenyingjie@pku.edu.cn,lewes6369@gmail.com}\\
}

\begin{document}

\maketitle
\vspace{-10mm}
\begin{figure*}[htb]
	\centering
	\includegraphics[width=\linewidth]{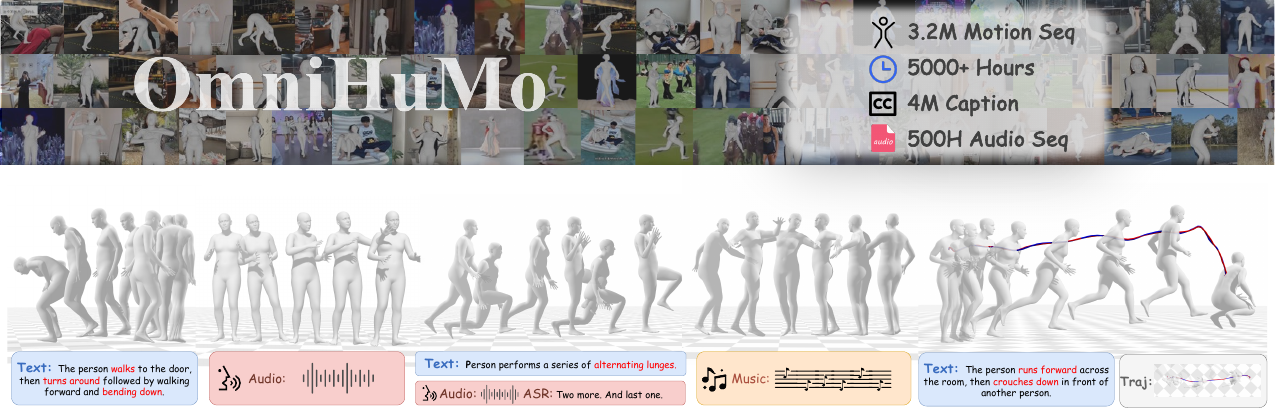}
    \caption{\textbf{Top}: OmniHuMo is a large-scale, high-quality human motion dataset with multimodal annotations. \textbf{Bottom}: We present AnyMo, a unified framework for controllable motion generation from diverse modalities and their combinations.}
	\label{teaser} 
\end{figure*}

\input{sections/0_abstract}
\input{sections/1_intro}

\input{sections/2_related_work}

\input{sections/6_data}

\input{sections/3_method}

\input{sections/4_exp}

\input{sections/5_conclusion}

\bibliographystyle{plain}
\bibliography{ref}
\input{sections/7_suppl}

\end{document}

%% file: sections/0_abstract.tex
\begin{abstract}

Conditional human motion generation remains a fundamental challenge in computer vision and robotics. Despite significant progress, current methods are often constrained by fixed modality configurations and task-specific architectures, leaving cross-modal interactions and the scaling laws of multimodal-conditioned synthesis largely underexplored. A key bottleneck is the scarcity of large-scale modality-aligned motion data, limiting generalization across diverse control signals.
In this work, we introduce \textbf{OmniHuMo}, a large-scale, high-quality dataset comprising over 5,000 hours of motion and 3.2 million sequences with precisely aligned multimodal annotations (e.g., text, speech, music, and trajectory). Leveraging OmniHuMo, we propose \textbf{AnyMo}, a unified multimodal framework combining a Residual FSQ-based motion tokenizer with a scalable masked modeling transformer, enabling  high-quality motion synthesis under arbitrary modality combinations. 
Extensive experiments show that AnyMo achieves high-fidelity synthesis while offering flexible control over both spatial and stylistic attributes. Code and models are available on our \href{https://interestingzhuo.github.io/anymo-page/}{\textcolor{blue}{project page}}. 
\end{abstract}
\vspace{-1mm}

%% file: sections/1_intro.tex
\section{Introduction}
\label{sec:intro}

Driven by the growing demands of digital media and robotics, human motion generation has advanced rapidly in recent years. The goal is to synthesize realistic and temporally coherent motions under various control signals. Recent progress in generative modeling has enabled motion generation from multiple modalities, including natural language descriptions \cite{zhang2023t2m, guo2024momask, cen2024generating_motion_in_scene, pinyoanuntapong2024mmm, tevet2022mdm, motionstreamer2024, rempe2026kimodo}, music \cite{tseng2023edge, siyao2022bailando, li2021aist}, speech \cite{chen2025lom, liu2024emage}, and spatial trajectories \cite{wan2024tlcontrol, xie2023omnicontrol, pinyoanuntapong2025maskcontrol}. 

Despite these advances, achieving precise controllability and robust cross-modal generalization remains challenging due to two key bottlenecks:
\textbf{First, Scarcity of large-scale, multimodal aligned motion data.} 
Most existing motion datasets \cite{guo2022humanml3d, mahmood2019amass, Plappert2016KIT, ionescu2013human3.6m} rely on an optical motion capture system, which provides high-fidelity sequences but is costly and labor-intensive, limiting scale and diversity.
Recent efforts \cite{zhang2025motionx++, zhang2025opendance, fan2025gotozero, wang2024scaling_motionlib}  explore motion extraction from in-the-wild videos, successfully scaling data to a million-level. Nevertheless,  these datasets lack comprehensive multimodal alignment (see Tab.~\ref{tab:data_statistic}), reducing their effectiveness for multimodal motion generation.
\textbf{Second, lack of a multimodally controllable generative framework.} 
Most existing methods focus on single-modality-driven motion generation \cite{guo2024momask, motionstreamer2024, li2021aist, liu2024emage, tseng2023edge, rempe2026kimodo}, limiting  flexibility in complex scenarios. 
Although recent works \cite{li2025genmo, chen2025lom, zhang2025motionanything} incorporate multiple modalities within a unified framework, they still follow a "multi-task but single-input" paradigm, treating each modality as an isolated generation task. As a result, they cannot explicitly model cross-modal dependencies required for simultaneous conditioning, and struggle to generalize to arbitrary combinations of control signals for a single motion sequence.

\begin{table}[t]
  \centering
  \small
  \caption{Comparison with existing motion datasets. "Mono MoCap" refers to markerless monocular video-based motion capture. "Data Agg" denotes datasets constructed by aggregating existing sources. }
  \resizebox{\linewidth}{!}{
  \begin{tabular}{l|ccc|cccc}
    \toprule
    Datasets & Clips & Duration & Source & Text & Speech & Music & RGB \\
    \midrule
    BEATv2 \cite{liu2024emage} & 1.8K & 60H & Marker MoCap & \ding{55} & \ding{51} & \ding{55} & \ding{55} \\
    HumanML3D \cite{guo2022humanml3d} & 14.6K & 28.6H & Marker MoCap & \ding{51} & \ding{55} & \ding{55} & \ding{55} \\
    OpenDanceSet \cite{zhang2025opendance} & 41K & 100.3H & Mono MoCap & \ding{55} & \ding{55} & \ding{51} & \ding{55} \\
    Motion-X++ \cite{zhang2025motionx++} & 120K & 181H & Data Agg \& Mono MoCap &\ding{51} & \ding{55} & \ding{51} & \ding{55} \\
    MotionHub \cite{ling2024versatilemotion} & 400K & 596H & Data Agg &\ding{51} & \ding{51} & \ding{51} & \ding{55} \\
    MotionMillion \cite{fan2025gotozero} & 2M & 2000H & Data Agg \& Mono MoCap & \ding{51} & \ding{55} & \ding{55} & \ding{55} \\
    \midrule
    OmniHuMo(Ours) & 3.2M & 5048H & Mono MoCap & \ding{51} & \ding{51} & \ding{51} & \ding{51} \\
    \bottomrule
  \end{tabular}}
  \label{tab:data_statistic}
  \vspace{-5mm}
\end{table}

Building on the above challenges, we argue that achieving robust generalization and flexible controllability in human motion generation requires both large-scale multimodal data and scalable model architectures. Accordingly, we identify two key components for any-modality motion generation: 1) a large-scale motion dataset with semantically aligned multimodal annotations, and 2) a scalable framework that supports motion synthesis under arbitrary combinations of input modalities.

To this end, we present \textbf{OmniHuMo}, a large-scale human motion dataset with rich multimodal annotations, constructed via an efficient pipeline for automatic labeling of web-scale videos. 
OmniHuMo offers three key advantages: 
1) \textbf{Large scale:} over 5,000 hours of motion data (3.2M+ sequences) extracted from web videos, as summarized in  Tab.~\ref{tab:data_statistic};
2) \textbf{Multimodal annotations:} all sequences are paired with textual descriptions, with a subset ($\sim$ 500 hours) further annotated with speech or music;
3) \textbf{High quality:} rigorous filtering and post-processing ensure reliable motion reconstruction and consistent annotations.

Building upon OmniHuMo, we propose \textbf{AnyMo}, a scalable masked modeling framework for motion generation under arbitrary combinations of conditioning signals. The architecture comprises two key components.
\textbf{1) R-FSQ-based Motion Tokenizer.}
While Finite Scalar Quantization (FSQ) \cite{mentzer2023fsq} improves stability and efficiency over Vector Quantization \cite{van2017vqvae}, it still suffers from information loss. To address this, we adopt residual quantization \cite{guo2024momask}, where a base stream captures coarse motion structure and subsequent streams progressively encode residual details, reducing reconstruction error.
\textbf{2) Scalable Masked Transformer.}
We employ a LLaMA-based \cite{touvron2023llama} Transformer backbone for motion token modeling. Unlike autoregressive methods,  AnyMo uses bidirectional attention with masked modeling \cite{guo2024momask} to capture global temporal dependencies across both past and future frames. For multimodal conditioning, modality-specific encoders process heterogeneous inputs, enabling flexible combinations of control signals. In addition,  Parallel Mask Modeling is designed to predict all residual streams simultaneously without sequence flattening, effectively improving efficiency and generation quality. 

To the best of our knowledge, OmniHuMo is the largest human motion dataset to date that integrates text, audio, and visual modalities. Extensive experiments demonstrate that AnyMo achieves competitive performance across diverse motion generation tasks, underscoring its superior effectiveness and cross-modal versatility.

%% file: sections/2_related_work.tex
\vspace{-3mm}
\section{Related Work}
\label{sec:related}
\vspace{-3mm}

In this section, we review related work on Motion Generation and Generative Masked Transformers. Related work on Data-driven Motion Modeling is provided in Appendix \ref{sec:related_work_more}.

\textbf{Motion Generation with Diverse Modalities.} \ 
Human motion generation has evolved from single-modality tasks---such as text-driven motion generation \cite{zhang2023t2m, guo2022humanml3d, rempe2026kimodo, guo2024momask, cao2026opent2m, wen2025hy-motion, tevet2022mdm, cen2024generating_motion_in_scene, li2025morph}, audio-conditioned gesture \cite{liu2024emage, liu2022disco, yi2023generating_TalkSHOW} and dance synthesis \cite{tseng2023edge, siyao2022bailando, li2021aist}
---toward more advanced paradigms that integrate multiple modalities \cite{li2025genmo, chen2025lom, zhang2025motionanything} to enhance controllability. 
However, due to the scarcity of well-aligned multimodal motion data, most existing methods adopt a shared backbone for multiple single-modality tasks, rather than explicitly modeling cross-modal interactions. Consequently, these approaches struggle to support arbitrary input modalities combinations. 
This lack of high-quality multimodal alignment remains a key bottleneck for scalable and flexible motion generation.

\textbf{Generative Masked Transformers.}  \ 
BERT \cite{devlin2018bert} introduces masked modeling for language, where a subset of tokens is randomly masked and a bidirectional Transformer is trained to reconstruct them. This paradigm has been successfully extended to other domains \cite{chang2022maskgit, zhu2025llada, you2025llada-v}.
In motion generation, Generative Masked Modeling (GMM) formulates synthesis as a non-autoregressive “mask-and-in-between” task, offering a favorable trade-off between quality and efficiency. Methods such as MoMask \cite{guo2024momask} and MMM \cite{pinyoanuntapong2024mmm} capture bidirectional temporal dependencies, while BAMM \cite{pinyoanuntapong2024bamm} further explores hybrid bidirectional–autoregressive modeling.
However, existing GMM approaches are primarily designed for single-modality control and typically model motion as a single-stream token sequence. Their extension to flexible multimodal conditioning and multi-stream token generation remains underexplored.  Scaling such frameworks to large datasets 
remains an open challenge.

%% file: sections/6_data.tex
\section{OmniHuMo Dataset}

\begin{figure}[t]
    \centering
    \includegraphics[width=0.9\linewidth]{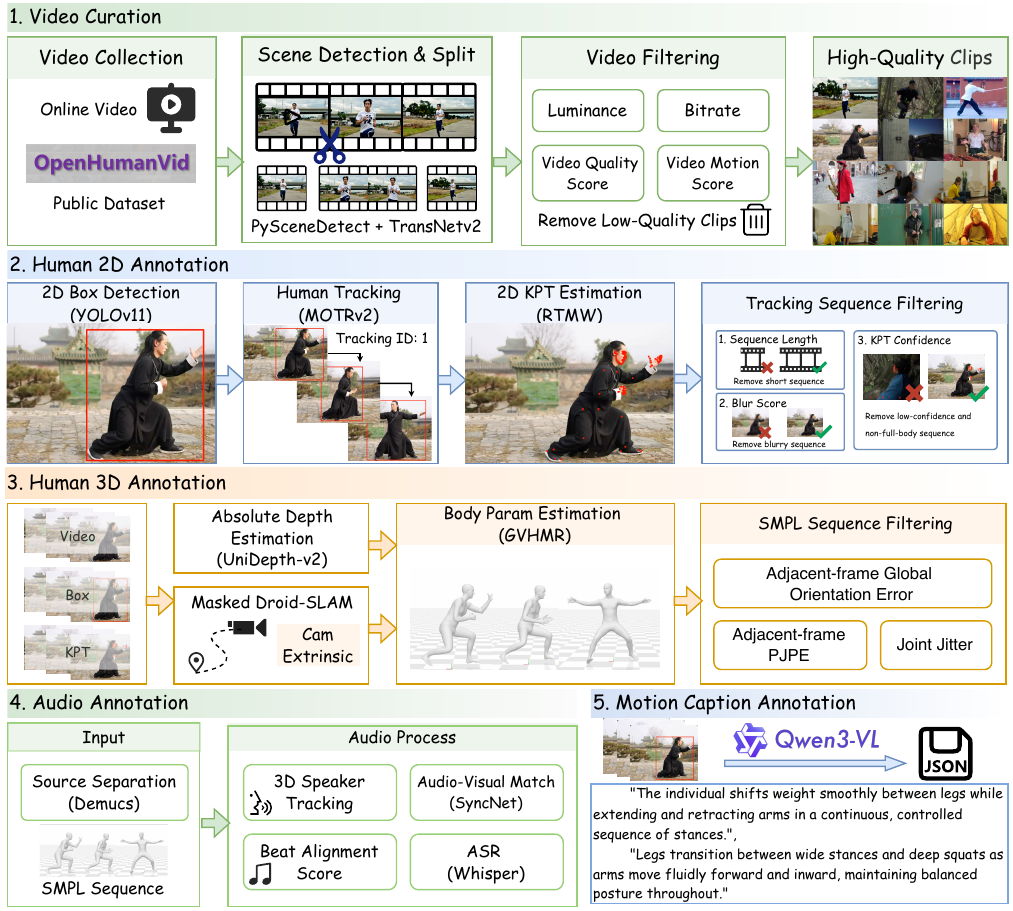}
    \caption{\textbf{Data Construction Framework of OmniHuMo.} The proposed pipeline systematically extracts high-quality human motion data with temporally aligned audio signals and corresponding textual descriptions.}
    \label{fig:data_framework}
    \vspace{-5mm}
\end{figure}

We introduce \textbf{OmniHuMo}, a large-scale \textbf{Omni}-modal \textbf{Hu}man \textbf{Mo}tion dataset, constructed entirely from diverse online videos. To enable scalable and diverse human motion data collection, we design an automated data collection pipeline, as illustrated in Fig.~\ref{fig:data_framework}. The pipeline consists of five sequential stages: Video Curation;  Human 2D Annotation; Human 3D Annotation; Audio Annotation; and Motion Caption Annotation. 

\subsection{Data Construction Pipeline}
\label{sec:data_construct_pipeline}

\noindent\textbf{Video Curation.} \ Video curation pipeline starts with large-scale video collection from online platforms and public datasets \cite{li2025openhumanvid}. This process yields over 200 million videos spanning diverse scenarios, actions, and recording conditions. 
To enhance the robustness of subsequent motion estimation, we perform scene detection and segmentation to mitigate artifacts caused by abrupt temporal transitions. We adopt a coarse-to-fine strategy: PySceneDetect~\cite{pyscenedetect} is first used to detect coarse scene boundaries based on pixel intensity and brightness variations, followed by TransNetV2~\cite{soucek2024transnet} to refine complex transitions such as fades and wipes. Videos are then segmented into single-scene clips.  Finally, we apply strict quality filtering based on heuristics including luminance, bitrate, visual quality, and motion intensity. Details about filtering criteria are provided in Appendix \ref{sec:Video Filtering}.

\textbf{Human 2D Annotation.} \ To reconstruct human motion in world coordinates, we first extract high-fidelity 2D annotations, including bounding boxes, keypoints, and tracking associations. Specifically, YOLOv11~\cite{khanam2024yolov11} is used for human detection, MOTRv2~\cite{zhang2023motrv2} for long-term human tracking.  Building upon these tracked sequences, 2D human poses are estimated using RTMW~\cite{jiang2024rtmw}. 
To ensure annotation quality and temporal consistency, we apply three filtering criteria: \textit{1) Temporal duration}: sequences shorter than 60 frames are discarded. 
\textit{2) Visual fidelity}: sequences with a mean blur score below 0.1 are removed to reduce degradation caused by motion blur. \textit{3) Pose reliability}: frames with an average keypoint confidence below 0.6 
are filtered out to reduce severe occlusions and incomplete poses. 

\textbf{Human 3D Annotation.} \ 
We employ GVHMR \cite{shen2024world_GVHMR} to reconstruct 3D human motion in world coordinates. Giving 2D bounding boxes, keypoints, video frames, and camera extrinsics,  GVHMR regresses SMPL \cite{SMPL:2015} parameters of 3D human motion sequences.    
To improve temporal consistency, we further filter out sequences with abrupt root orientation changes or excessive joint jitter caused by camera motion. Detailed criteria is provided in Appendix \ref{sec:Human 3D Annotation}.

\textbf{Audio Annotation.} \ 
Audio is an important modality for multimodal motion generation, especially for speech-driven gestures and music-driven dance. We first extract audio tracks from videos and use Demucs~\cite{rouard2023hybrid_Demucs} to separate vocals and background music.
\textbf{For dance video identification}, we compute the Beat Alignment Score (BAS)  between the music track and the SMPL motion sequence. Samples with BAS above 0.15 are classified as dance sequences, ensuring strong motion-music synchronization.
\textbf{For speech videos identification}, we adopt a three-stage pipeline: 3D-Speaker~\cite{chen2025_3D-Speaker} is used for speaker tracking, SyncNet~\cite{chung2016SyncNet} for audio-visual synchronization, and Whisper~\cite{radford2023whisper} for speech transcriptions. Samples without valid linguistic content are removed.

\textbf{Motion Caption Annotation.} \ 
To support semantic understanding and controllable motion synthesis, we generate textual captions for each motion sequence. Source videos are first segmented into  clips based on reconstructed motion trajectories. We then use Qwen3-VL-32B~\cite{Qwen3-VL}
to generate fine-grained action descriptions, with prompts focusing on localized human bounding boxes to capture detailed motion dynamics. 
Detailed prompt configurations are provided in Appendix \ref{sec:caption_annotation}.

\vspace{-2mm}
\subsection{Dataset Statistics}
\vspace{-2mm}

\begin{figure}[!b]
  \centering
  \begin{minipage}{0.32\textwidth}
    \centering
    \includegraphics[height=2.5cm]{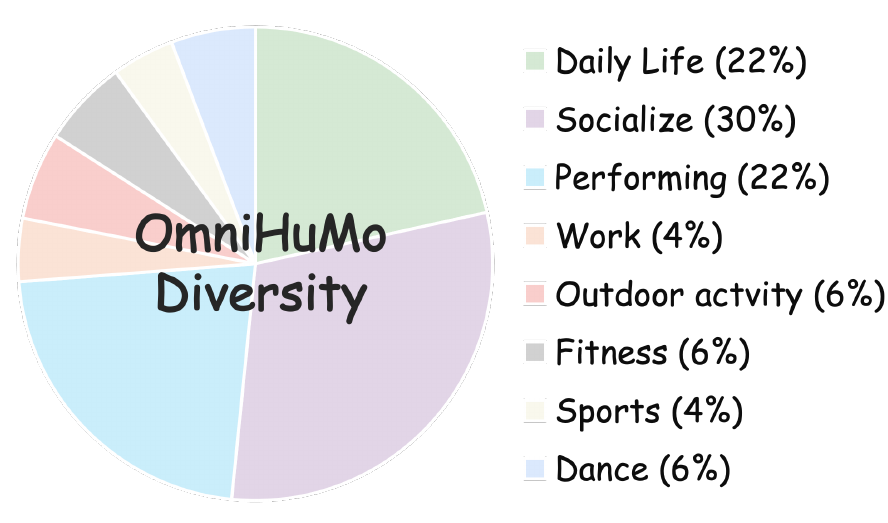}
    \caption{OmniHuMo diversity.}
    \label{fig:motion_diversity}
  \end{minipage}
  \hfill
  \begin{minipage}{0.31\textwidth}
    \centering
    \includegraphics[height=2.5cm]{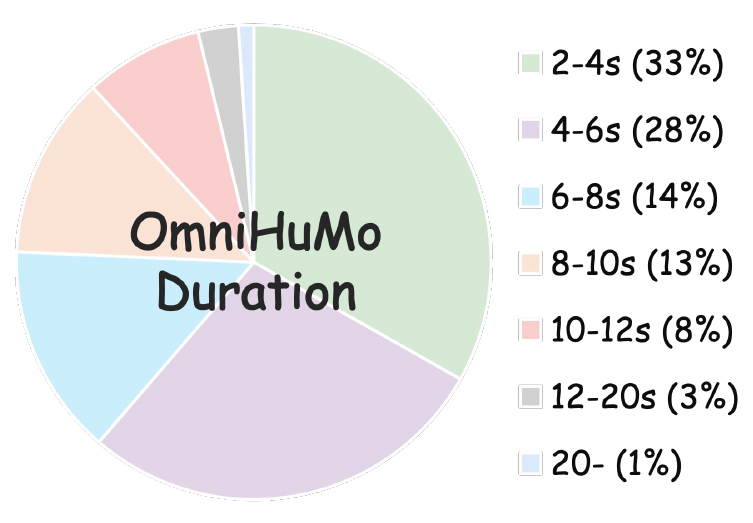}
    \caption{OmniHuMo duration.}
    \label{fig:motion_duration1}
  \end{minipage}
  \hfill
  \begin{minipage}{0.33\textwidth}
    \centering
    \includegraphics[height=2.5cm]{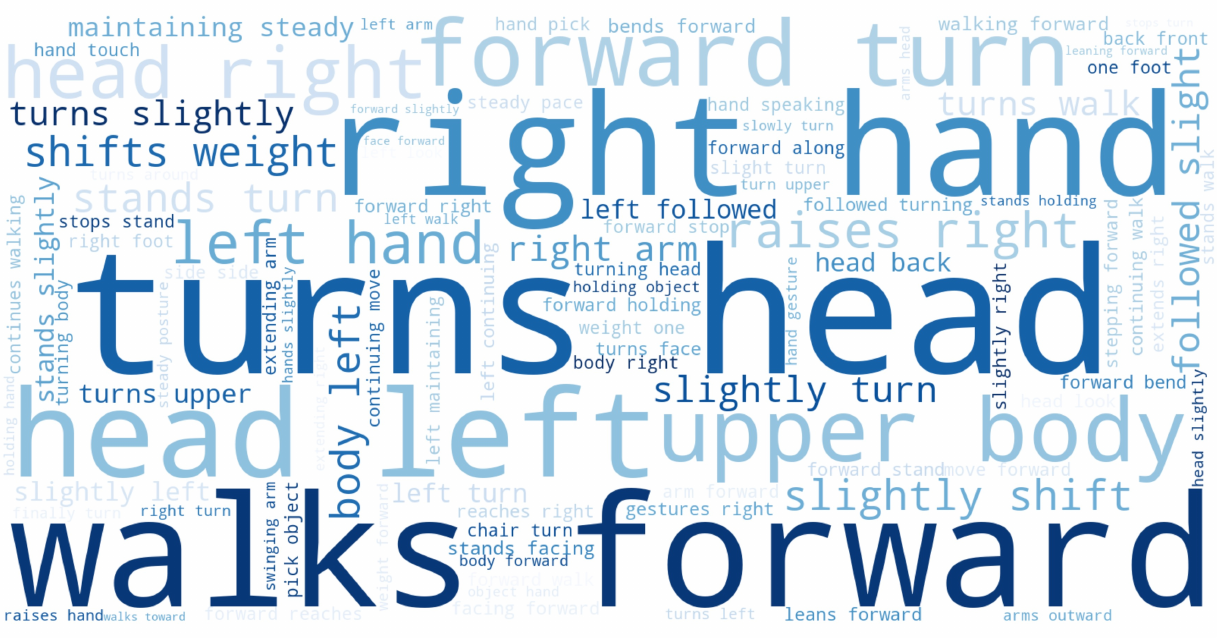}
    \caption{Word cloud of Caption.}
    \label{fig:motion_duration2}
  \end{minipage}
  \vspace{-5mm}
\end{figure}

As summarized in Tab.~\ref{tab:data_statistic}, OmniHuMo contains over \textbf{5,000 hours} of human motion data and more than \textbf{3.2 million motion sequences}. Due to the heterogeneous video sources, modality annotations are unevenly distributed. Each sequence is paired with 1–3 textual captions, while a subset (approximately 500 hours) additionally contains temporally aligned audio. This design balances large-scale motion diversity with high-quality multimodal annotations, supporting both general motion synthesis and audio-driven generation.

We further analyze motion categories and sequence durations in Fig.~\ref{fig:motion_diversity}, Fig.~\ref{fig:motion_duration1} and Fig.~\ref{fig:motion_duration2}. OmniHuMo covers diverse activities, from indoor scenarios (e.g., artistic performances and choreographed dance) to outdoor activities (e.g., sports and daily events). Most Sequence are 2--10 seconds long, making them suitable for modeling atomic actions and rapid motion transitions.

%% file: sections/3_method.tex
\section{Method}

We propose \textbf{AnyMo}, a scalable masked modeling framework for 3D motion generation conditioned on arbitrary combinations of text, music, speech, and trajectory inputs. Trained on the large-scale OmniHuMo dataset, AnyMo generates motion sequences $\mathbf{X} \in \mathbb{R}^{T \times D}$ guided by these multimodal conditions, where $T$ denotes sequence length and $D$ the pose dimensionality. 

\begin{figure}[t]
\centering
\begin{subfigure}[t]{0.37\linewidth}
\centering
\includegraphics[width=\linewidth]{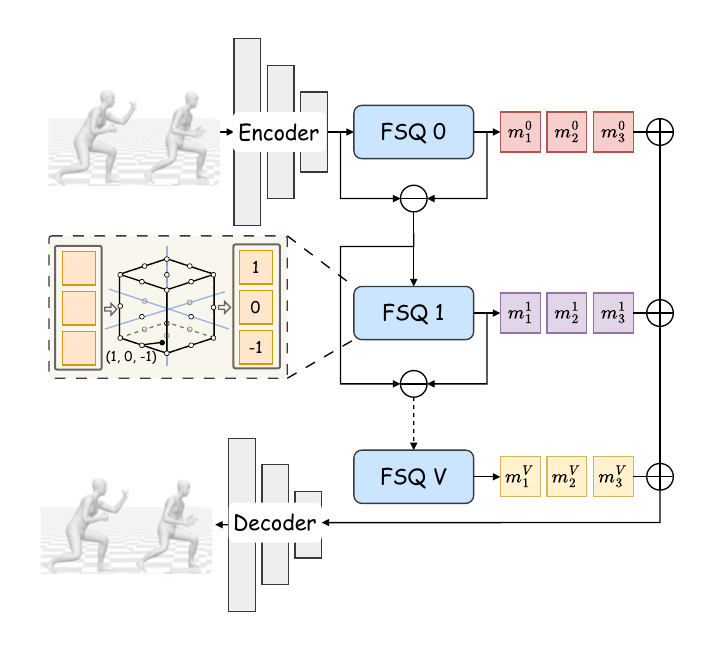}
\caption{Residual FSQ.}
\label{fig:res_fsq}
\end{subfigure}
\hfill
\begin{subfigure}[t]{0.62\linewidth}
\centering
\includegraphics[width=\linewidth]{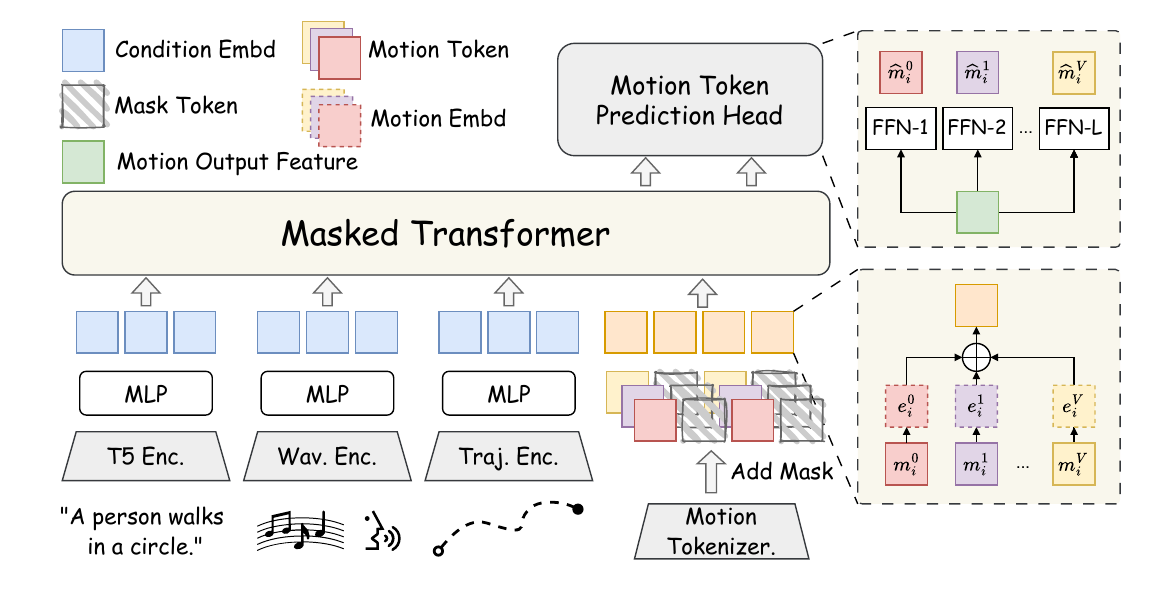}
\caption{Scalable masked transformer architecture.}
\label{fig:model_framework}
\end{subfigure}
\caption{\textbf{Overview of AnyMo}. The framework consists of two components. First, we train a motion tokenizer based on Residual FSQ to discretize continuous motion into multi-stream discrete tokens. Second, we train a masked Transformer that supports diverse conditioning signals, including text, audio, and trajectories, as well as their combinations, to generate coherent human motion sequences.}
\label{fig:framework_and_fsq}
\end{figure}

As illustrated in Fig.~\ref{fig:framework_and_fsq}, AnyMo consists of two key components: 1) a  Residual Finite Scalar Quantizer-based tokenizer that discretizes continuous motion into  hierarchical tokens (Sec. \ref{sec:r-fsq}); and 2) a scalable masked Transformer built upon LLaMA \cite{touvron2023llama} architecture, which reconstructs motion from masked motion tokens (Sec. \ref{src:transformer}).

\subsection{Motion Residual FSQ-VAE}\label{sec:r-fsq}
Conventional discrete motion representations typically rely on vector quantization with a single, capacity-limited codebook, which suffers from two key limitations. First, the non-differentiable quantization nearest-neighbor assignment can lead to codebook collapse, leading to highly imbalanced code usage and inefficient utilization of the latent space. Second, single-stage quantization compresses complex motion patterns into a coarse representation, limiting its ability to capture fine-grained dynamics.
To address these issues, we adopt  \textbf{Residual Finite Scalar Quantization (R-FSQ)}, which replaces vector matching by deterministic scalar discretization in a bounded space with hierarchical quantization. This design promotes uniform code utilization and enables progressive multi-scale refinement, leading to high-fidelity reconstruction.

As shown in Fig. \ref{fig:res_fsq},  R-FSQ comprises a motion encoder $E$, decoder $D$, and $V+1$ hierarchical quantization stages.  
Given a motion sequence $\mathbf{X}\in\mathbb{R}^{T\times D}$, the encoder $E$ maps it to a continuous latent representation $\mathbf{Z}\in\mathbb{R}^{t\times d}$, where $T/t$ denotes the temporal downsampling ratio. To avoid codebook collapse in conventional VQ, each stage employs a Finite Scalar Quantizer (FSQ). For each vector $\boldsymbol{z}\in\mathbb{R}^d$ within $\mathbf{Z}$, FSQ first applies a bounding function $f\left(\cdot\right)$ (instantiated as $\operatorname{sigmoid}$) to constrain its value within a predefined range. Each dimension $z_i$ is then discretized into $L_i$ levels:
\begin{equation}
\hat{z}_i = \mathrm{FSQ}\left(z_i\right) = \mathrm{round}\left(f\left(z_i\right) \cdot \left(L_i - 1\right)\right), ~~ \mathrm{for} ~i = 0, \ldots, d-1.
\end{equation}
This operation maps $\boldsymbol{z}$ onto a discrete coordinate $\left(\hat{z}_0,\dots,\hat{z}_{d-1}\right)$ on a regular grid. The $d$-dimensional coordinate can be flattened into a scale via a bijective mapping, yielding a codebook with $|\mathcal{C}| = \prod_{i=0}^{d-1} L_i$ discrete codes.

To capture fine-grained motion dynamics, R-FSQ performs recursive residual quantization over  $V+1$ FSQ stages. Let $\mathbf{R}^0=\mathbf{Z}$, the model computes a quantized approximation $\mathbf{\widehat{Z}}^v$ and updates the residual at each stage $v$:
\begin{equation}
\mathbf{\widehat{Z}}^v = \mathrm{FSQ}\left(\mathbf{R}^v\right),~~\mathbf{R}^{v+1}=\mathbf{R}^v-\mathbf{\widehat{Z}}^v,~~ \mathrm{for} ~v = 0, \ldots, V.
\end{equation}
The final quantized representation is obtained by aggregating all hierarchical approximations: $\mathbf{\widehat{Z}}=\sum_{v=0}^V \mathbf{\widehat{Z}}^v$. R-FSQ is optimized end-to-end via a reconstruction objective:
\begin{equation}
\mathcal{L}_{1} = \left\|\mathbf{X}-D\left(\mathbf{\widehat{Z}}\right)\right\|_2^2.
\end{equation}

 The R-FSQ converts the motion sequence $\mathbf{X}$ into $V + 1$ ordered discrete token sequences $\{\boldsymbol{m}^0, \boldsymbol{m}^1, \ldots, \boldsymbol{m}^V\}$, where $\boldsymbol{m}^v \in \left\{0,\dots,|\mathcal{C}|-1\right\}^{t}$. This yields a coarse-to-fine discrete representation: $\boldsymbol{m}^0$ encodes global motion patterns, while higher levels refine fine-grained dynamics.

\subsection{Scalable Masked Transformer}
\label{src:transformer}

\noindent\textbf{Multi-modal Condition Encoders.} \
As shown in Fig.~\ref{fig:model_framework}, we use modality-specific encoders to process heterogeneous inputs, including text, audio, and motion trajectories. We adopt the pre-trained T5-XL encoder \cite{chung2024scaling_t5} to extract semantic representations from text. For audio, we use the encoder from  WavTokenizer \cite{ji2024wavtokenizer} to capture features from both speech and music domains. For motion trajectories,  we employ a lightweight convolutional encoder to capture local temporal dependencies. 
All modality-specific features are projected into a common $c$-dimensional embedding space for unified conditioning. Formally, the encoded features are represented as $Z_{\mathrm{text}} \in \mathbb{R}^{N_{t} \times c}$, $Z_{\mathrm{audio}} \in \mathbb{R}^{N_{a} \times c}$ and $Z_{\mathrm{traj}} \in \mathbb{R}^{N_{tr} \times c}$, where $N_{t}, N_{a}$ and $N_{tr}$ denote the sequence length of respective modality.

\noindent\textbf{Parallel Mask modeling.} \ 
Following the masked motion modeling paradigm in MoMask~\cite{guo2024momask}, we formulate motion generation as a masked token reconstruction task, where the model learns to recover original motion sequences from partially corrupted inputs. To enable scalable generation, we adopt a LLaMA-based bidirectional Transformer backbone, which supports global context reasoning beyond conventional autoregressive approaches. 
Built upon R-FSQ, each motion sequence $\mathbf{X}$ is represented as $V+1$ parallel streams of discrete tokens $\{\boldsymbol{m}^0, \boldsymbol{m}^1, \ldots, \boldsymbol{m}^V\}$. This multi-stream representation introduces the challenge of jointly modeling hierarchical residual structure. To this end, we propose \textbf{parallel masked modeling strategy} that enables simultaneous encoding and prediction across all token streams, as shown in  Fig.~\ref{fig:model_framework}.

Specifically, we implement a \textit{consistent} masking scheme across the $V+1$ streams by randomly replacing a subset of tokens with a special $\mathrm{<MASK>}$ token. Under this scheme, once a temporal index is selected, the tokens at that timestep across  all residual levels  are masked simultaneously.   Let $\widetilde{\boldsymbol{m}}^v$ denote the masked token sequence at the $v$-th residual level. 
To process these multi-stream tokens in parallel, we introduce a set of independent embedding layers $\left\{\mathrm{Embd}^v\left(\cdot\right)\right\}_{v=0}^{V}$ that map each $\widetilde{\boldsymbol{m}}^v$ into a shared latent space. The comprehensive motion representation is then obtained by aggregating embeddings across all streams: 
\begin{equation}
Z_{\mathrm{enc}} = \sum_{v=0}^{V}\mathrm{Embd}^{v}\left(\widetilde{\boldsymbol{m}}^v\right),
\end{equation}
which is subsequently fed into the Transformer backbone for contextual modeling.

For parallel token prediction, we employ $V+1$ prediction heads $\left\{\mathrm{FFN}^v\left(\cdot\right) \right\}_{v=0}^{V}$, each parameterized by an independent feed-forward network. Given the latent representation $h$ produced from the transformer, each head reconstructs its respective token stream as $\widehat{\boldsymbol{m}}^v=\mathrm{FFN}^v\left(h\right)$.

\noindent \textbf{Objective Function.} \ The over Masked Transformer is optimized via a cross-entropy loss applied independently to each token stream:
\begin{equation}
\mathcal{L}_{2} = -\sum_{v=0}^{V}\mathrm{log}~p\left(\boldsymbol{m}^{v}|Z_{\mathrm{text}}, Z_{\mathrm{audio}}, Z_{\mathrm{traj}}, \widetilde{\boldsymbol{m}}^{v}\right)
\end{equation}

\subsection{Training Paradigm}

Conventional motion generation frameworks typically rely on strictly paired datasets for training. However, in large-scale data collection, motion sequences are often not perfectly synchronized with audio signals (e.g., music or speech), as data collection prioritizes scene diversity and motion variability. As a result, high-quality audio–motion aligned data account for only about one-tenth of the OmniHuMo Dataset ($\sim$ 500 hours). To address this limitation, we adopt a staged training curriculum to enable robust cross-modal alignment under weakly aligned data.

\noindent\textbf{Stage I: Text-to-motion Pre-training.} \ 
We first train the model on text-to-motion task using all text-motion pairs in OmniHuMo. The audio and trajectory encoders are frozen. This stage learns a strong motion prior and aligns textual features with a structured motion representation space, providing a stable foundation for subsequent multimodal integration.

\noindent\textbf{Stage II: Multi-modal Alignment.} \
We then introduce audio-aligned data while freezing the text encoder and Transformer backbone, updating only the audio and trajectory encoders. This stage maps audio and trajectory features into the text-aligned latent space, enabling a unified cross-modal representation. 

\noindent\textbf{Stage III: Joint Multi-Modal Fine-tuning.} \ 
Finally, we fine-tune the full model for arbitrary multimodal conditioning. To address 
 the imbalance between text-only data and audio-motion aligned data, we use disproportional sampling: 10\% of text-only data and full audio-aligned subset are used per epoch. We further apply modality augmentation by randomly injecting trajectory input for text-conditioned samples and textual input for audio-conditioned samples with probability 0.1. This improves robustness and supports flexible multimodal conditioning.

%% file: sections/4_exp.tex
\section{Experiments}
\label{sec:exper}

\begin{table}[t]
\small
  \centering
  \begin{minipage}[t]{0.35\linewidth}
      \centering
      \caption{Reconstruction performance under different data scales. }
      \begin{tabular}{c|cc}
        \toprule
        Data Size & FID $\downarrow$ & MPJPE $\downarrow$ \\
        \midrule
        0.05M & 160.65 & 94.55 \\
        0.6M & 43.18 & 44.65 \\
        3M & \textbf{17.32} & \textbf{27.92} \\
        \bottomrule
      \end{tabular}
  \label{tab:data_size_ablation_fsq}
  \end{minipage}
  \hfill
  \begin{minipage}[t]{0.59\linewidth}
      \centering
      \caption{Ablation of different motion token modeling strategies. AR means auto-regressive prediction.}
      \resizebox{\linewidth}{!}{
          \begin{tabular}{cc|ccc}
            \toprule
            Strategy & Methods & FID $\downarrow$ & R@1 $\uparrow$ & MMDist $\downarrow$ \\
            \midrule
            A & AR-Flatten & 26.71 & 0.62 & 17.17 \\
            B & Mask-Flatten & 20.89 & 0.66 & \textbf{16.54}  \\
            C & Mask-Parallel & \textbf{19.46} & \textbf{0.66} & 16.78  \\
            \bottomrule
          \end{tabular}
      }
      \label{tab:motion_decode_stragety}
  \end{minipage}
    \vspace{-3mm}
\end{table}

\begin{table}[t]
  \centering
  \small
  \caption{Motion reconstruction performance comparison measured in MPJPE (mm). }
  \begin{tabular}{l|l|ccc}
    \toprule
    \multirow{2}{*}{Method} & \multirow{2}{*}{Train Dataset} & \multicolumn{3}{c}{Test Dataset} \\ \cline{3-5}
    & & HumanML3D \cite{guo2022humanml3d} & MotionMillion \cite{fan2025gotozero} & OmniHuMo \\
    \midrule
    ScaMo \cite{lu2025scamo} & MotionUnion \cite{lu2025scamo} &  63.3 & 88.9 & - \\
    GoToZero \cite{fan2025gotozero} & MotionMillion \cite{fan2025gotozero} & 41.9 & 45.5 & 36.1 \\
    Ours (R-FSQ) & OmniHuMo & \textbf{27.9} & \textbf{21.5} & \textbf{13.2} \\
    \bottomrule
  \end{tabular}
  \label{tab:recon_cross_dataset}
  \vspace{-3mm}
\end{table}

We summarize the experimental results in this section. Due to space limitations, additional implementation details and evaluation metrics are provided in Appendix~\ref{sec:exp_setup}, with more ablation study in Appendix~\ref{sec:albation_study_appendix} and qualitative results in Appendix ~\ref{sec:qualizative_res}.

\subsection{Experimental Setup}

\noindent\textbf{Datasets.} \ 
We conduct experiments on HumanML3D \cite{guo2022humanml3d}, and our OmniHuMo dataset. OmniHuMo provides semantically aligned multimodal annotations. For the text-conditioned subset (\textbf{OmniHuMo-Text}), we split the data into 50K test, 10K validation, and the rest for training. For speech and music modalities (\textbf{OmniHuMo-Speech} and \textbf{OmniHuMo-Music}), we use  8K test, 2K validation, and the rest for training. 
For ablation studies, we use HumanML3D \cite{guo2022humanml3d}, a widely used 3D human motion-language benchmark. It contains 14,616 motion clips and 44,970 text descriptions (28.59 hours), with each clip annotated by 3–4 captions. We follow the standard 80\%/5\%/15\% train/validation/test split.

\noindent\textbf{Implementation Details.} \ 
The motion tokenizer adopts a 4-layer Residual FSQ with a codebook size of 2048 per layer. The encoder–decoder follows SnapMoGen \cite{snapmogen2025}, combining convolutional residual blocks and self-attention \cite{vaswani2017attention}, with a temporal downsampling factor of 4. It is trained with a learning rate of 2e-4 for 200 epochs on 16 NVIDIA H20 GPUs, with a batch size of 256 per GPU. 
The AnyMo network is built upon LLaMA \cite{touvron2023llama}. To study scaling laws with respect to model capacity, we train models ranging from 111M to 3B parameters. AnyMo is trained for 210 epochs with an initial learning rate of $2\times10^{-4}$, decayed to $1\times10^{-5}$ after 500 warm-up steps using cosine scheduling. Training is performed on 48 NVIDIA H20 GPUs with a batch size of 16 per GPU.

\noindent\textbf{Evaluation Metrics.} \  
For motion reconstruction, we use Mean Per Joint Position Error (MPJPE) to measure geometric accuracy. 
For text-driven motion generation, following T2M-GPT \cite{zhang2023t2m}, we report Motion–Text Retrieval Precision (R-Precision), Fréchet Inception Distance (FID), Multimodal Distance (MMDist), and Motion Diversity (Div).
For speech-driven gesture and music-driven dance generation, following LoM \cite{chen2025lom}, we use FID, Beat Alignment Score (BAS), and Div. 

\subsection{Ablation Study}
\label{sec:albation_study}

\noindent\textbf{Data scale for motion tokenizer.} \ We study the effect of data scale on R-FSQ tokenizer.  For reconstruction, the tokenizer is trained on 
OmniHuMo and evaluated on HumanML3D.  As shown in Tab.~\ref{tab:data_size_ablation_fsq}, reconstruction performance consistently improves as the training data scale increases, demonstrating the importance of large-scale data for learning accurate motion representations. 

\noindent\textbf{Motion token modeling strategy. }
We compare different motion token modeling strategies in Tab.~\ref{tab:motion_decode_stragety}.
Both training and evaluation are conducted on HumanML3D. First, comparing Strategy A and B, masked modeling outperforms autoregressive modeling, highlighting the benefit of bidirectional context. Second, comparing Strategy B and C, the parallel modeling strategy further improves generation quality. Moreover, compared to the flattening strategy, the parallel design enables simultaneous decoding of multiple token streams, improving computational efficiency.

\begin{table}[!t]
  \centering
  \small
  \caption{Comparison of text-driven motion generation performance on OmniHuMo-Text test set.}
  \begin{tabular}{l|cccccc}
    \toprule
    Method & FID $\downarrow$ & R@1 $\uparrow$ & R@2 $\uparrow$ & R@3 $\uparrow$ & MMDist $\downarrow$ & Div $\rightarrow$ \\
    \midrule
    Real & - & 0.74 & 0.88 & 0.93 & 25.75 & 46.59 \\
    \midrule
    AnyMo-111M & 262.10 & 0.63 & 0.76 & 0.82 & 29.16 & 43.57 \\
    AnyMo-343M & 216.26 & 0.67 & 0.80 & 0.86 & 28.23 & 44.98 \\
    AnyMo-775M & 148.81 & 0.71 & 0.83 & 0.88 & 27.24 & 45.36 \\
    AnyMo-1B & 102.21 & 0.74 & 0.86 & 0.90 & 26.28 & 45.74 \\
    AnyMo-3B & \textbf{55.59} & \textbf{0.75} & \textbf{0.87} & \textbf{0.91} & \textbf{25.87} & \textbf{46.71} \\
    \bottomrule
  \end{tabular}
  \label{tab:t2m}
  \vspace{-2mm}
\end{table}

\begin{table}[!t]
  \centering
  \small
  \begin{minipage}[t]{0.48\textwidth}
    \centering
    \caption{Comparison of speech-driven motion generation performance on  OmniHuMo-Speech test set.}
    \label{tab:speech2m}
    \begin{tabular}{l|ccc}
      \toprule
      Method & FID $\downarrow$ & BAS $\uparrow$ & Div $\rightarrow$ \\
      \midrule
      Real & - & 0.205 & 44.40 \\
      \midrule
      AnyMo-111M & 178.83 & 0.204 & 42.54 \\
      AnyMo-343M & 201.01 & 0.202 & 42.00 \\
      AnyMo-775M & \textbf{83.80} & 0.205 & 42.86 \\
      AnyMo-1B & 96.87 & 0.208 & 42.66 \\
      AnyMo-3B & 91.12 & \textbf{0.214} & \textbf{43.47} \\
      \bottomrule
    \end{tabular}
  \end{minipage}
  \hfill
  \begin{minipage}[t]{0.48\textwidth}
    \centering
    \caption{Comparison of music-driven motion generation performance on  OmniHuMo-Music test set.}
    \label{tab:music2m}
    \begin{tabular}{l|ccc}
      \toprule
      Method & FID $\downarrow$ & BAS $\uparrow$ & Div $\rightarrow$ \\
      \midrule
      Real & - & 0.210 & 39.22 \\
      \midrule
      AnyMo-111M & 70.98 & 0.207 & 39.78 \\
      AnyMo-343M & 74.18 & 0.209 & \textbf{39.13} \\
      AnyMo-775M & \textbf{34.41} & 0.210 & 38.22 \\
      AnyMo-1B & 37.62 & 0.211 & 36.99 \\
      AnyMo-3B & 46.17 & \textbf{0.213} & 38.15 \\
      \bottomrule
    \end{tabular}
  \end{minipage}
  \vspace{-2mm}
\end{table}

\subsection{Motion Reconstruction Comparison}

We compare motion reconstruction performance of tokenizers trained on different datasets in Tab.~\ref{tab:recon_cross_dataset}, we evaluate three methods: our R-FSQ tokenizer trained on OmniHuMo, ScaMo's FSQ \cite{lu2025scamo} trained on MotionUnion, and GoToZero's tokenizer \cite{fan2025gotozero} trained on MotionMillion. All models are evaluated on the test sets of HumanML3D \cite{guo2022humanml3d}, MotionMillion \cite{fan2025gotozero}, and OmniHuMo.  As shown in Tab.~\ref{tab:recon_cross_dataset}, our method consistently achieves the best reconstruction performance across all benchmarks. We attribute this gain to the larger scale and diversity of OmniHuMo, as well as the residual quantization design to better preserve fine-grained motion details.

\subsection{Single-Modality Motion Generation Performance}

\noindent\textbf{Text-driven Motion Generation.} \ 
Following prior work \cite{petrovich2022temos}, we train a motion-text retrieval model on the OmniHuMo-Text training set and evaluate AnyMo on its test split. As shown in Tab.~\ref{tab:t2m}, performance improves consistently as the model scales from 111M to 3B parameters, indicating stable gains under the current data and training regime, consistent with empirical scaling laws.

\noindent\textbf{Audio-driven Motion Generation.} \ 
We evaluate \textit{Speech-to-Gesture} and \textit{Music-to-Dance} on OmniHuMo-Speech and OmniHuMo-Music test sets, respectively. As shown in Tab. \ref{tab:speech2m} and Tab. \ref{tab:music2m}, audio-driven generation does not exhibit monotonic scaling behavior. While AnyMo-775M achieves the best FID, further increasing model capacity leads to slight performance degradation, suggesting potential overfitting under limited paired audio–motion data. In contrast, Beat Alignment Score (BAS) improves with model size, indicating stronger temporal modeling and better cross-modal synchronization between audio rhythm and motion.

\begin{table}[!t]
  \centering
  \small
  \caption{Comparison of multi-modal conditional inputs in text-driven motion generation task. Evaluated on OmniHuMo-Text test set using AnyMo-3B.}
  \begin{tabular}{cc|ccccc}
    \toprule
    Text & Traj & FID $\downarrow$ & R@1 $\uparrow$ & \makecell{Trajectory Error\\(>50cm)~(\%)~$\downarrow$} & \makecell{Location Error\\(>50cm)~(\%)~$\downarrow$} & \makecell{Average Error\\(cm)~$\downarrow$}\\
    \midrule
    \ding{52} & & 55.59 & 0.75 & 0.52 & 0.28 & 50.50 \\
    \ding{52} & \ding{52} & \textbf{41.43} & \textbf{0.77} & \textbf{0.33} & \textbf{0.14} & \textbf{27.16}\\
    \bottomrule
  \end{tabular}
  \label{tab:multi-cond-on-text-driven-gen}
\end{table}

\begin{table}[!t]
  \centering
  \small
  \begin{minipage}[t]{0.49\linewidth}
      \centering
      \caption{Comparison of multi-modal conditional inputs in speech-driven motion generation task.}
      \resizebox{\linewidth}{!}{
          \begin{tabular}{ccc|ccc}
            \toprule
            Speech & Text & Traj & FID $\downarrow$ & BAS $\uparrow$ & Div $\rightarrow$ \\
            \midrule
            \ding{52}(Real) & & & - & 0.205 & 44.40 \\
            \midrule
            \ding{52} &  &  & 91.12 & 0.214 & 43.37 \\
            \ding{52} & \ding{52} &  & 89.74 & 0.215 & \textbf{43.63} \\
            \ding{52} &  & \ding{52} & 76.82 & 0.217 & 42.51 \\
            \ding{52} & \ding{52} & \ding{52} & \textbf{76.55} & \textbf{0.217} & 43.18 \\
            \bottomrule
          \end{tabular}
      } 
      \label{tab:multi-cond-on-audio-driven-gen}
  \end{minipage}
  \hfill
  \begin{minipage}[t]{0.49\linewidth}
      \centering
      \caption{Comparison of multi-modal conditional inputs in music-driven motion generation task. }
      \resizebox{\linewidth}{!}{
          \begin{tabular}{ccc|ccc}
            \toprule
            Music & Text & Traj & FID $\downarrow$ & BAS $\uparrow$ & Div $\rightarrow$ \\
            \midrule
            \ding{52}(Real) & & & - & 0.210 & 39.22 \\
            \midrule
            \ding{52} & & & 46.17 & 0.213 & \textbf{38.15} \\
            \ding{52} & \ding{52} &  & 43.84 & 0.213 & 37.34 \\
            \ding{52} & & \ding{52} & \textbf{42.99} & 0.214 & 37.93 \\
            \ding{52} & \ding{52} & \ding{52} & 43.26 & \textbf{0.215} & 36.94 \\
            \bottomrule
          \end{tabular}
    }
      \label{tab:multi-cond-on-music-driven-gen}
  \end{minipage}
\end{table}

\subsection{Multi-Modality Motion Generation Performance}

\textbf{Text-driven Motion Generation.} \ Tab. \ref{tab:multi-cond-on-text-driven-gen} reports results under multi-modal conditioning for text-driven motion generation. Incorporating auxiliary modalities consistently improves performance across metrics. In particular, adding trajectory information enhances motion realism (FID), retrieval accuracy (R@1), and reduces trajectory error. These results indicate that trajectory cues provide useful structural priors for improving both motion quality and text alignment.

\textbf{Audio-driven Motion Generation.} \ Tab. \ref{tab:multi-cond-on-audio-driven-gen} and Tab. \ref{tab:multi-cond-on-music-driven-gen}  present results for speech- and music-driven motion generation under multi-modal conditioning. Overall, incorporating additional modalities improves performance. Compared to text, trajectory conditioning yields larger gains in motion realism (FID) and beat alignment (BAS), highlighting stronger structural guidance. However, motion diversity does not consistently increase with more modalities, likely due to the reduced flexibility of the motion space under additional conditioning constraints.

%% file: sections/5_conclusion.tex
\section{Conclusion}
\label{sec:conclusion}

In this paper, we introduce \textbf{OmniHuMo}, the first large-scale human motion dataset with rich multimodal annotations, providing a foundation for multimodal motion modeling. Based on OmniHuMo, we propose \textbf{AnyMo}, a unified framework for controllable motion generation from arbitrary combinations of text, speech, music, and trajectories. Experiments show that our method achieves high-quality and diverse motion synthesis with flexible multimodal control. These results highlight the importance of large-scale, well-aligned data for improving generalization and controllability. We hope OmniHuMo and AnyMo will advance general-purpose human motion generation and multimodal generative modeling.

%% file: sections/7_suppl.tex
\newpage
\appendix
\setcounter{figure}{0}
\setcounter{table}{0}
\renewcommand{\thefigure}{S\arabic{figure}}
\renewcommand{\thetable}{S\arabic{table}}

\section*{\centering{Appendix}} \label{appendix}

This Appendix is orginazed into the following sections: Section~\ref{sec:related_work_more} present additional related work on Data-driven Motion Modeling. Section~\ref{sec:details_of_data_cons} provide details on the construction of the OmniHuMo dataset. Section~\ref{sec:exp_setup} describe the experimental setup, including implementation details and evaluation metrics. Section~\ref{sec:albation_study_appendix} present additional ablation studies on the R-FSQ tokenizer. Section~\ref{sec:qualizative_res} show visualization examples of AnyMo. Section~\ref{sec:limitation} discuss the limitation and future work of our work.

\section{Related Work}
\label{sec:related_work_more}
\textbf{Data-driven Motion Modeling.} \ 
Motion synthesis quality heavily depends on the scale and diversity of training data. Traditional optical motion capture datasets, such as AMASS \cite{mahmood2019amass} and HumanML3D \cite{guo2022humanml3d}, are constrained by high acquisition costs and limited variety. Recent efforts have explored automatic motion extraction from in-the-wild videos. For example, MotionX++ \cite{zhang2025motionx++} proposes an automated pipeline that leverages pose estimation techniques  \cite{cai2023smplerx} to extract motion sequences from large-scale internet videos. Building on this paradigm, \cite{fan2025gotozero, wang2024scaling_motionlib, cao2026opent2m, wen2025hy-motion} introduce million-scale motion datasets, marking significant progress toward large-scale motion modeling.
However, these datasets primarily focus on text-conditioned generation and lack comprehensive multimodal annotations. To address this gap, we propose an efficient pipeline for harvesting motion with precisely aligned multimodal annotations from web videos, enabling scalable multimodal motion generation.

\section{Details of OmniHuMo Construction}\label{sec:details_of_data_cons}

\begin{figure}[htbp]
    \centering
    \includegraphics[width=0.85\linewidth]{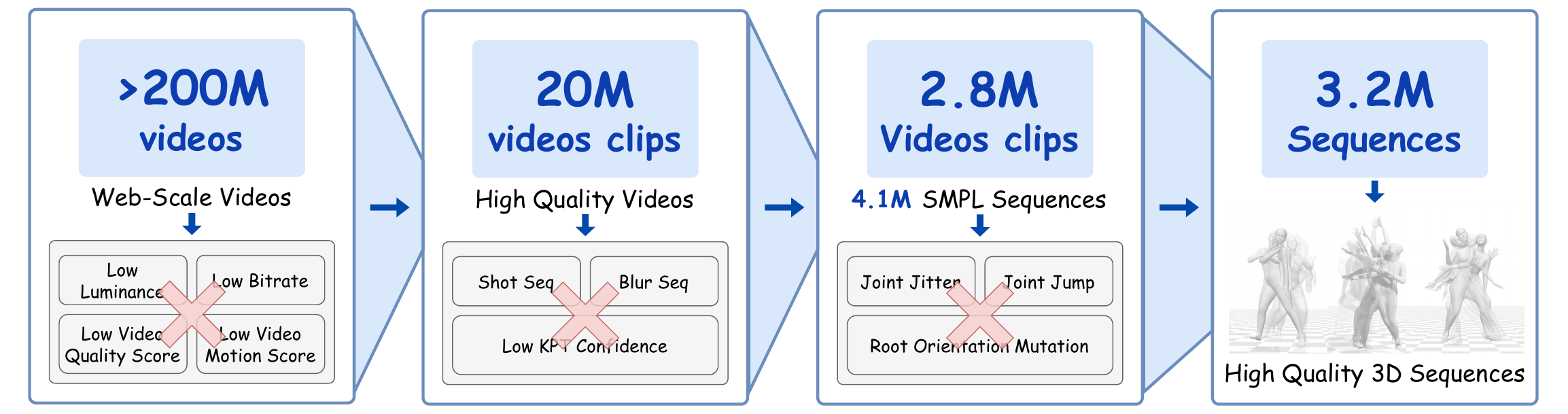}
    \caption{Filter operators in the data processing pipeline. }
    \label{fig:data_filter}
\end{figure}

This section describes the filtering procedures for raw videos and 3D motion annotations, as illustrated in Fig.~\ref{fig:data_filter}. We apply strict filtering criteria, resulting in 3.2M motion sequences distilled from over 200M raw videos. We also present the prompts used for data captioning. 

\begin{figure}[t]
    \centering
    \includegraphics[width=\linewidth]{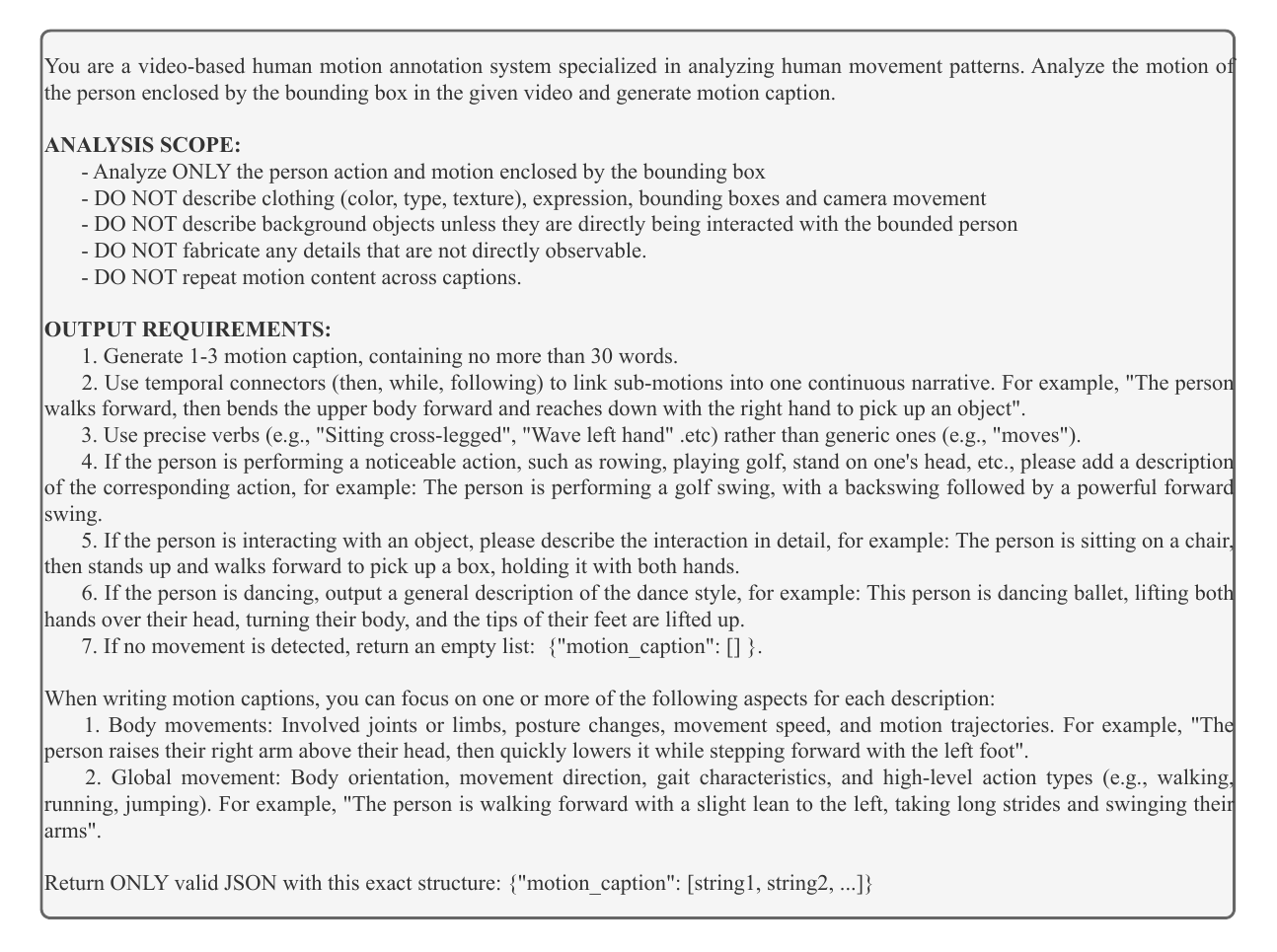}
    \caption{Prompt design for text description annotation in OmniHuMo.}
    \label{fig:prompt}
\end{figure}

\begin{figure}[t]
    \centering
    \includegraphics[width=0.85\linewidth]{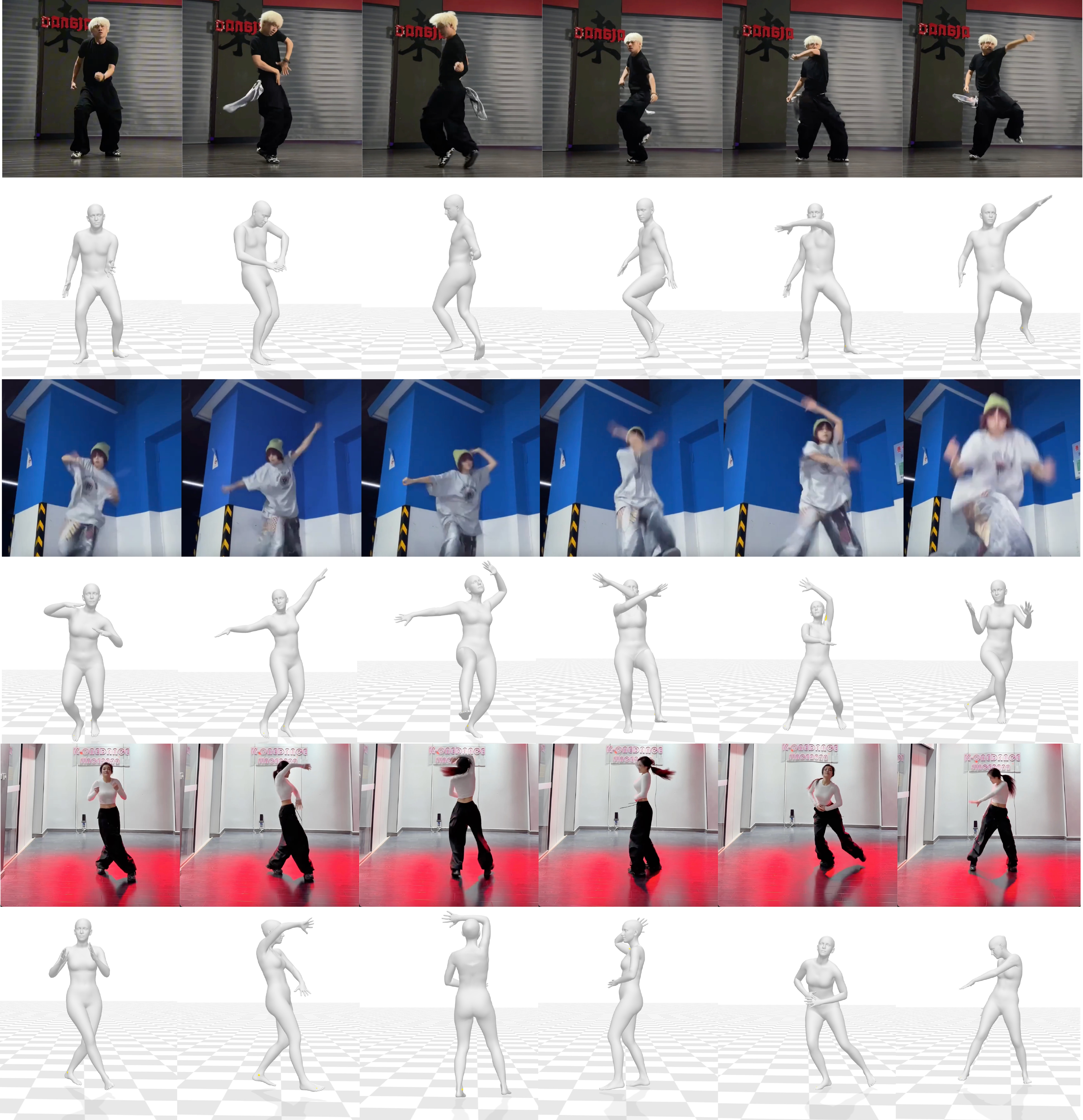}
    \caption{Visualization of 3D SMPL reconstructions for motion sequences in OmniHuMo. }
    \label{fig:data_vis_smpl}
\end{figure}

\subsection{Video Filtering. }
\label{sec:Video Filtering}

The collected source videos contain a substantial amount of low-quality content. To ensure reliable downstream annotation, we apply a series of filtering criteria.

\textbf{Average bitrate}. \ Bitrate reflects information density and serves as a proxy for visual quality. We compute the normalized average bitrate as $B / \sqrt{W \times H}$, where $W$, $H$ and $B$ denote resolution and bitrate. Videos with a normalized bitrate below 500 are discarded.

\textbf{Luminance.} \ Following \cite{li2025openhumanvid}, we measure overall video brightness to filter extreme lighting conditions. The luminance score is computed as $0.2126R + 0.7152G + 0.0722B$, where $R$, $G$, and $B$ denote RGB values. Videos with luminance outside $\left[10, 210\right]$ are removed.

\textbf{Video quality score.} \ We first apply frame-level CLIP~\cite{clip} aesthetic scoring for efficient pre-filtering, removing videos with an average score below 4.0.  We then adopt DOVER \cite{wu2023dover} for comprehensive video quality assessment combining aesthetic and technical cues. Videos with a final score below 0.25 are filtered out. 


\textbf{Motion score.} \ High-motion scenes often introduce severe motion blur that degrades annotation accuracy, while overly static scenes provide limited informative motion cues. To address this, we estimate optical flow using UniMatch \cite{xu2023unimatch} and compute the average flow magnitude as a motion score. Videos with scores outside $\left[3.5, 350\right]$ are discarded.

\subsection{Human 3D Annotation}
\label{sec:Human 3D Annotation}

We employ GVHMR \cite{shen2024world_GVHMR} to reconstruct 3D human motion in world coordinates. It estimates poses in a Gravity-View (GV) coordinate system defined by gravity and camera viewing directions, which reduces ambiguity in world coordinate definition. The model takes bounding boxes, 2D keypoints, video frames, and relative camera rotations as inputs, and predicts SMPL parameters, including root translation $t$, body pose $\theta$, root rotation $r$, and shape parameters $\beta$.

To obtain camera motion, we employ DROID-SLAM \cite{teed2021droid} to estimate camera extrinsics. 
Since dynamic human regions degrade SLAM accuracy, we follow prior work to mask moving humans during Dense Bundle Adjustment. In practice, SAM2~\cite{ravi2024sam2} masks are unstable under fast motion. We instead use RF-DETR~\cite{rf-detr} detection boxes to mask dynamic regions, which improves the stability and accuracy of camera estimation.


To reduce instability in 3D reconstruction caused by occlusions and camera estimation errors, we further filter reconstructed SMPL sequences using the following criteria.

\textbf{Root Orientation Mutation.} \ Although GVHMR is robust, abrupt root orientation changes may still occur under rapid camera motion or occlusion. We measure the rotation difference  between consecutive frames:
\begin{equation}
    \Delta \theta_i = \arccos\!\left(\frac{\mathrm{tr}(R_i R_{i-1}^{\top}) - 1}{2}\right),
\end{equation}
where $R_i$ denotes the root rotation at frame $i$. Sequences with $\Delta \theta_i > 30^\circ$ are discarded.

\textbf{Joint Jitter.} \ We quantify temporal smoothness using jerk. For joint $j$ at frame $i$ with position $\mathbf{p}_{i,j}$, we compute:
\begin{equation}
    \mathbf{\dddot{J}}_{i,j} = \mathbf{p}_{i+1,j} - 3\mathbf{p}_{i,j} + 3\mathbf{p}_{i-1,j} - \mathbf{p}_{i-2,j},
\end{equation}
The average jerk over the sequence is:
\begin{equation}
    \mathbf{\dddot{J}} = \frac{1}{(T-3)J}\sum_{i=3}^{T-1}\sum_{j=1}^{J} \lVert \dddot{J}_{i,j} \rVert_2.
\end{equation}
Sequences with $\mathbf{\dddot{J}} > 0.015$ are discarded.

\textbf{Joint Position Jump.} \ To detect sudden local motion  anomalies, we compute the maximum frame-wise joint displacement:
\begin{equation}
\mathrm{Jump}_i = \max_{j} \left\| \mathbf{p}_{i,j} - \mathbf{p}_{i-1,j} \right\|_2.
\end{equation}
Sequences with $\mathrm{Jump}_i$ above 200mm are removed.

\subsection{Motion Caption Annotation}
\label{sec:caption_annotation}

We use Qwen3-VL-32B \cite{Qwen3-VL} to generate fine-grained action descriptions within detected human bounding boxes. The prompt design is shown in Fig.~\ref{fig:prompt}. To ensure quality, we constrain the model to describe only the target person’s actions and poses, excluding irrelevant information such as clothing, facial attributes, background, camera motion, or other unobservable details. Each motion sequence is annotated with 1–3 captions, each limited to 30 words. Captions are required to use precise action verbs (e.g., “standing on right leg,” “swinging left arm”) and temporal connectors (e.g., “then,” “while”) to form coherent action descriptions. For specific categories such as dance and sports, we additionally encourage explicit activity labels (e.g., “lat pulldown,” “Latin dance”) alongside the description.

\subsection{Visualization Examples}

\begin{figure}[!t]
    \centering
    \includegraphics[width=0.85\linewidth]{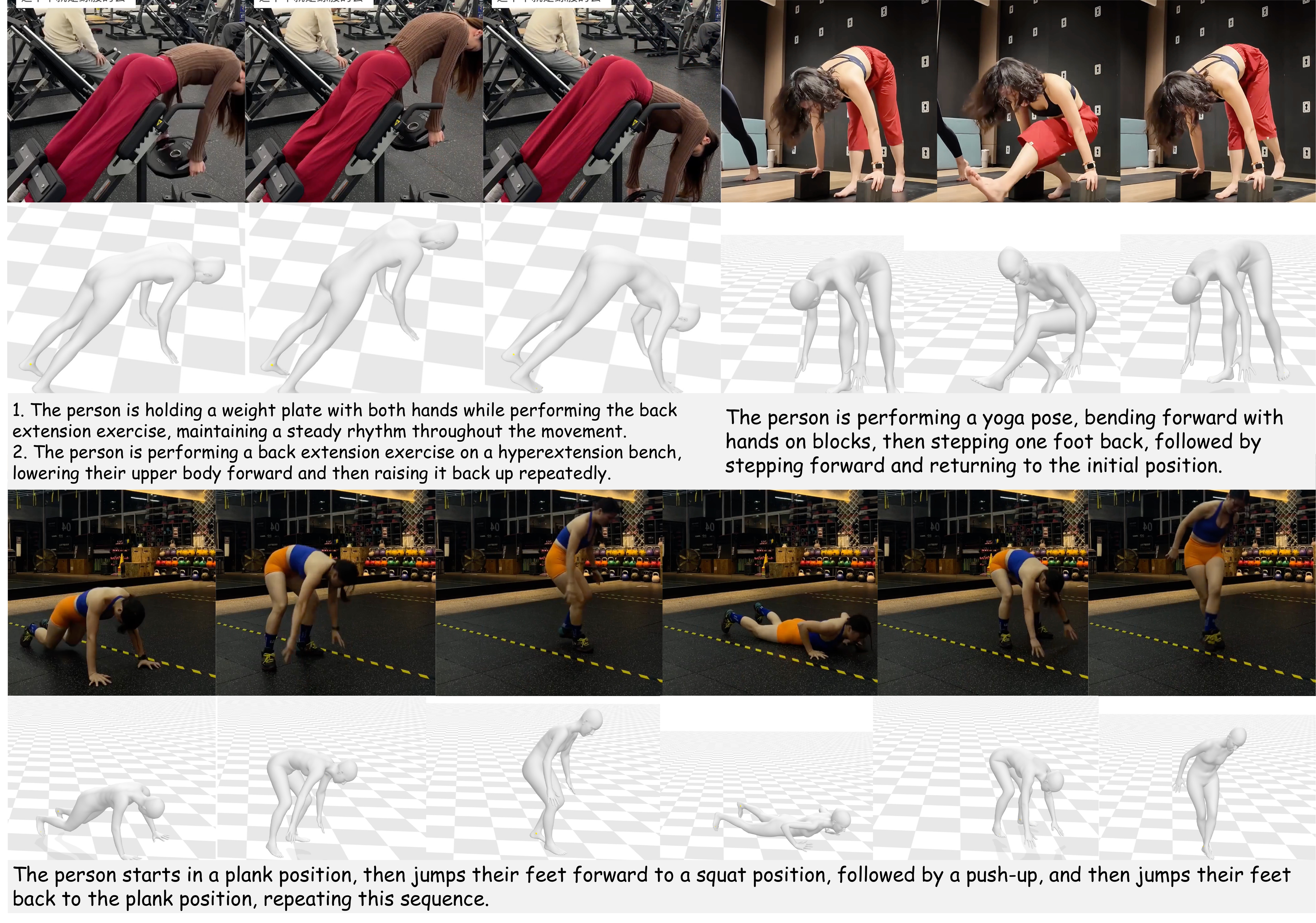}
    \caption{Visualization of SMPL reconstruction and the corresponding text description.}
    \label{fig:data_vis_caption}
\end{figure}

\begin{figure}[!t]
    \centering
    \includegraphics[width=0.85\linewidth]{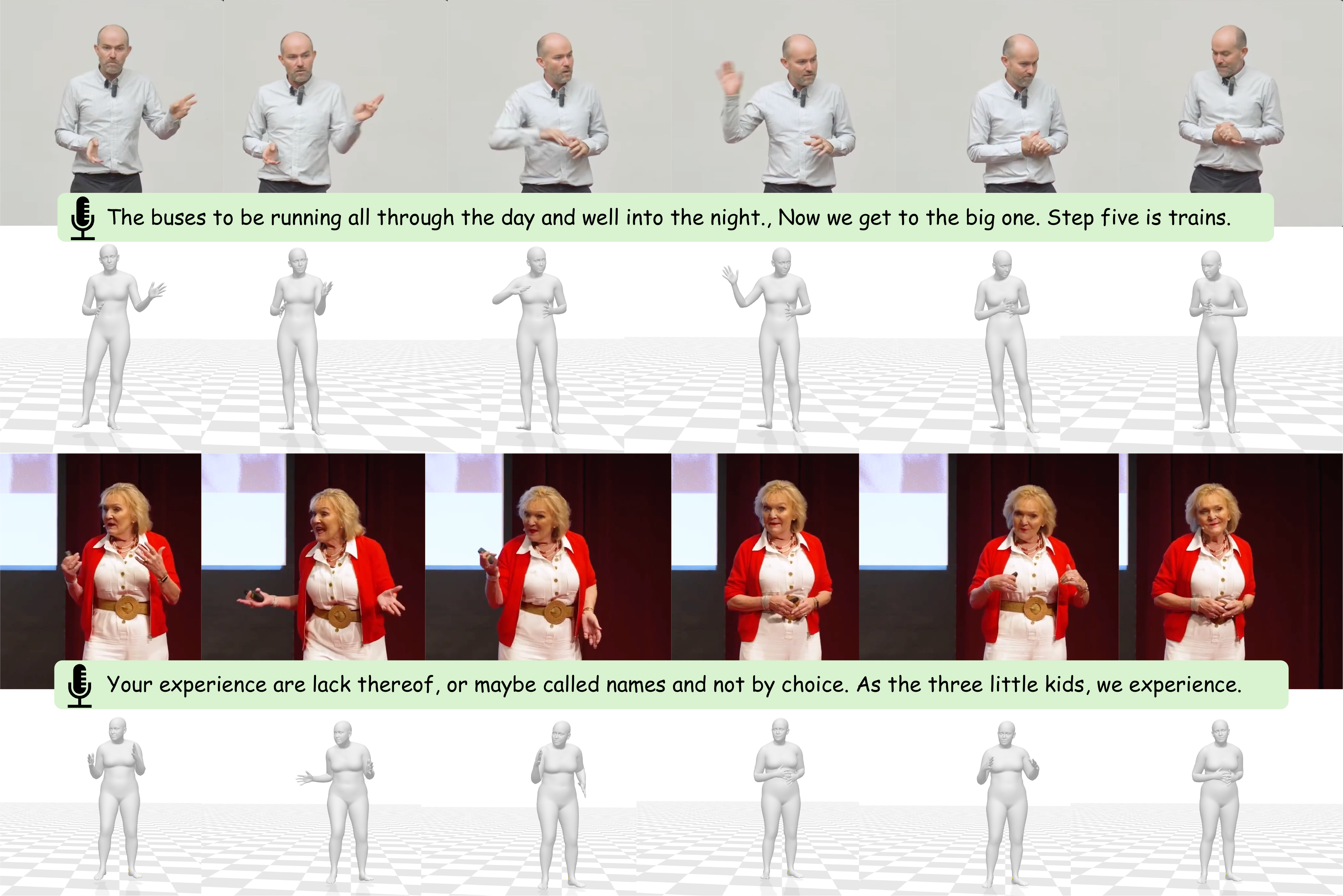}
    \caption{Visualization of the speaker's SMPL reconstruction and the corresponding transcript.}
    \label{fig:data_vis_audio}
\end{figure}

We present visualization examples of OmniHuMo in Fig.~\ref{fig:data_vis_smpl}, \ref{fig:data_vis_caption}, and \ref{fig:data_vis_audio}. These results demonstrate that OmniHuMo covers diverse motion patterns with strong multimodal alignment, providing a high-quality foundation for large-scale motion modeling.

\section{Experimental Setup}\label{sec:exp_setup}

\subsection{Implementation Details}

\textbf{Data Pipeline.} \ The data construction pipeline is deployed across multiple clusters. Video curation runs on a CPU cluster, while the Human 2D \& 3D and audio annotation are processed  on 100 L20 GPUs. Motion captioning is performed on a separate 40 H20 GPUs cluster. Overall, the pipeline generates approximately 100k high-quality motion sequences per day.

\textbf{Motion Tokenizer.} \ The motion tokenizer adopts a residual FSQ architecture with 4 layers and a codebook size of 2048 per layer.  The encoder and decoder follows SnapMoGen \cite{snapmogen2025}, consisting of alternating convolutional residual blocks and self-attention~\cite{vaswani2017attention}, with temporal downsampling by a factor of 4. It is trained on OmniHuMo using AdamW with an initial learning rate of $2\times10^{-4}$. A multi-step decay is applied at epochs $\left[60, 140\right]$ with a factor of 0.3, for 200 epochs. Training uses 16 NVIDIA H20 GPUs with a batch size of 256 per GPU. 

\textbf{AnyMo Training.} \ The AnyMo network is built upon the LLaMA architecture \cite{touvron2023llama}, with RMSNorm applied before attention and feed-forward layers. To study scaling behavior, we train models ranging from 111M to 3B parameters. Optimization uses AdamW with an initial learning rate of $2\times10^{-4}$, 500 warm-up steps, and cosine decay to $1\times10^{-5}$. Training runs for 210 epochs on 48 NVIDIA H20 GPUs with a batch size of 16 per GPU.

\subsection{Evaluation Metrics.}

\textbf{Motion Reconstruction}. \ We use Mean Per Joint Position Error (MPJPE) to measure geometric accuracy, computed as the average L2 distance between reconstructed and ground-truth joint positions across all frames.

\textbf{Text-driven Motion Generation}. \ Following T2M-GPT \cite{zhang2023t2m}, we evaluate text-driven motion generating using FID, R-Precision, Div, and MMDist:
\begin{itemize}[leftmargin=2em]
    \item \textbf{FID}: Fréchet Inception Distance measures the distribution gap between generated and real motions, computed as Fréchet distance between their feature distributions in embedding space.
    \item \textbf{R-Precision}: Motion–Text Retrieval Precision measures the alignment between generated motions and input text via retrieval accuracy. For each motion, we rank its Euclidean distances to 32 candidate text descriptions (1 ground truth and 31 randomly ssampled negatives) and report Top 1/2/3 retrieval accuracy. 
    \item \textbf{Div}: Diversity is computed as the average pairwise Euclidean distance between randomly sampled motion features, reflecting the spread of generated samples.
    \item \textbf{MMDist}: MultiModel Distance measures the average Euclidean distance between motion features and their corresponding text feature, indicating cross-modal alignment quality.
\end{itemize}

\textbf{Speech-driven Gesture Generation}.\  Following LoM \cite{chen2025lom}, we use FID, BAS and Div to evaluate gesture generation performance. FID and Div are computed in the same way as in text-driven motion generation. Beat Alignment Score (BAS) measures temporal synchronization between audio beats and generated motion beats. It is computed as the average Gaussian-weighted alignment between each audio beat and its nearest motion beat based on squared temporal distance.

\textbf{Music-driven Dance Generation}. \ We use the same evaluation metrics as in speech-driven gesture generation, including FID, BAS, and Div. Traditional dance generation \cite{siyao2022bailando} uses five metrics, including $\mathrm{FID}_k$ and $\mathrm{Div}_k$ for kinematic feature distribution and diversity, $\mathrm{FID}_g$ and $\mathrm{Div}_g$ for geometric feature distribution, and BAS for motion–music synchronization. However, EDGE \cite{tseng2023edge} shows that these kinematic and geometric metrics are unreliable due to heuristic feature design that fails to capture high-level semantics. Therefore, we adopt a contrastive learning-based feature extractor and report FID, BAS, and Div.

\section{Ablation Study}
\label{sec:albation_study_appendix}

This section presents additional ablation studies on the motion tokenizer  R-FSQ. For the reconstruction task, the motion tokenizer is trained on a 200K subset of OmniHuMo and evaluated on HumanML3D \cite{guo2022humanml3d}. For the motion generation task, both training and evaluation are conducted on HumanML3D.

\noindent\textbf{Codebook size on R-FSQ.} \ 
We study the effect of codebook size on the reconstruction performance, as shown in Tab.~\ref{tab:codebook_size_ablation}. The results indicate that reconstruction and generation quality improve consistently with larger codebooks. Notably, increasing training data yields more significant gains than scaling the codebook size alone.

\noindent\textbf{Number of residual layers on R-FSQ.} \ 
Tab.~\ref{tab:residual_layer_ablation} provides an ablation study on the number of residual layers, evaluated on both reconstruction and generation performance.
As shown in Tab.~\ref{tab:residual_layer_ablation}, reconstruction quality improves with increasing residual depth. However, generation performance degrades when the number of residual layers exceeds 4, likely due to increased prediction difficulty introduced by additional token streams that must be modeled jointly. 

\begin{table}[!htbp]
  \centering
  \label{tab:speech_music_comparison}
  \begin{minipage}[t]{0.49\linewidth}
    \centering
    \caption{Comparison of reconstruction and generation performance under different data scales. }
    \resizebox{\linewidth}{!}{
    \begin{tabular}{c|cc|cc}
        \toprule
        \multirow{2}{*}{\makecell{Codebook \\ Size}} & \multicolumn{2}{c|}{Reconstruction} & \multicolumn{2}{c}{Generation (T2M)}\\ \cline{2-5}
        & FID $\downarrow$ & MPJPE $\downarrow$ & FID $\downarrow$ & R@1 $\uparrow$ \\
        \midrule
        1024 & 101.86 & 76.05 & 34.67 & 0.59 \\
        2048 & 77.35 & 69.69 & 30.80 & 0.60 \\
        4096 & 72.24 & 65.71 & 30.26 & 0.60 \\
        8192 & 73.30 & 65.38 & 31.36 & 0.59 \\
        16384 & \textbf{71.63} & \textbf{64.93} & \textbf{28.52} & \textbf{0.60} \\
        \bottomrule
      \end{tabular}}
      \label{tab:codebook_size_ablation}
  \end{minipage}
  \hfill
  \begin{minipage}[t]{0.50\linewidth}
    \centering
    \caption{Comparison of reconstruction and generation performance under different residual depths.}
    \resizebox{0.95\linewidth}{!}{
    \begin{tabular}{c|cc|cc}
        \toprule
        \multirow{2}{*}{\makecell{Residual \\ Depth}} & \multicolumn{2}{c|}{Reconstruction} & \multicolumn{2}{c}{Generation (T2M)}\\ \cline{2-5}
        & FID $\downarrow$ & MPJPE $\downarrow$ & FID $\downarrow$ & R@1 $\uparrow$ \\
        \midrule
        1 & 101.86 & 76.05 & 33.01 & 0.61 \\
        2 & 81.64 & 61.54 & 25.32 & 0.62 \\
        4 & 52.80 & 49.06 & \textbf{19.46} & \textbf{0.66} \\
        6 & 45.90 & 41.47 & 19.93 & 0.65 \\
        8 & \textbf{38.58} & \textbf{34.79} & 23.64 & 0.60\\
        \bottomrule
  \end{tabular}}
  \label{tab:residual_layer_ablation}
  \end{minipage}
\end{table}

\section{Qualitative Results}\label{sec:qualizative_res}

\begin{figure}[t]
    \centering
    \includegraphics[width=\linewidth]{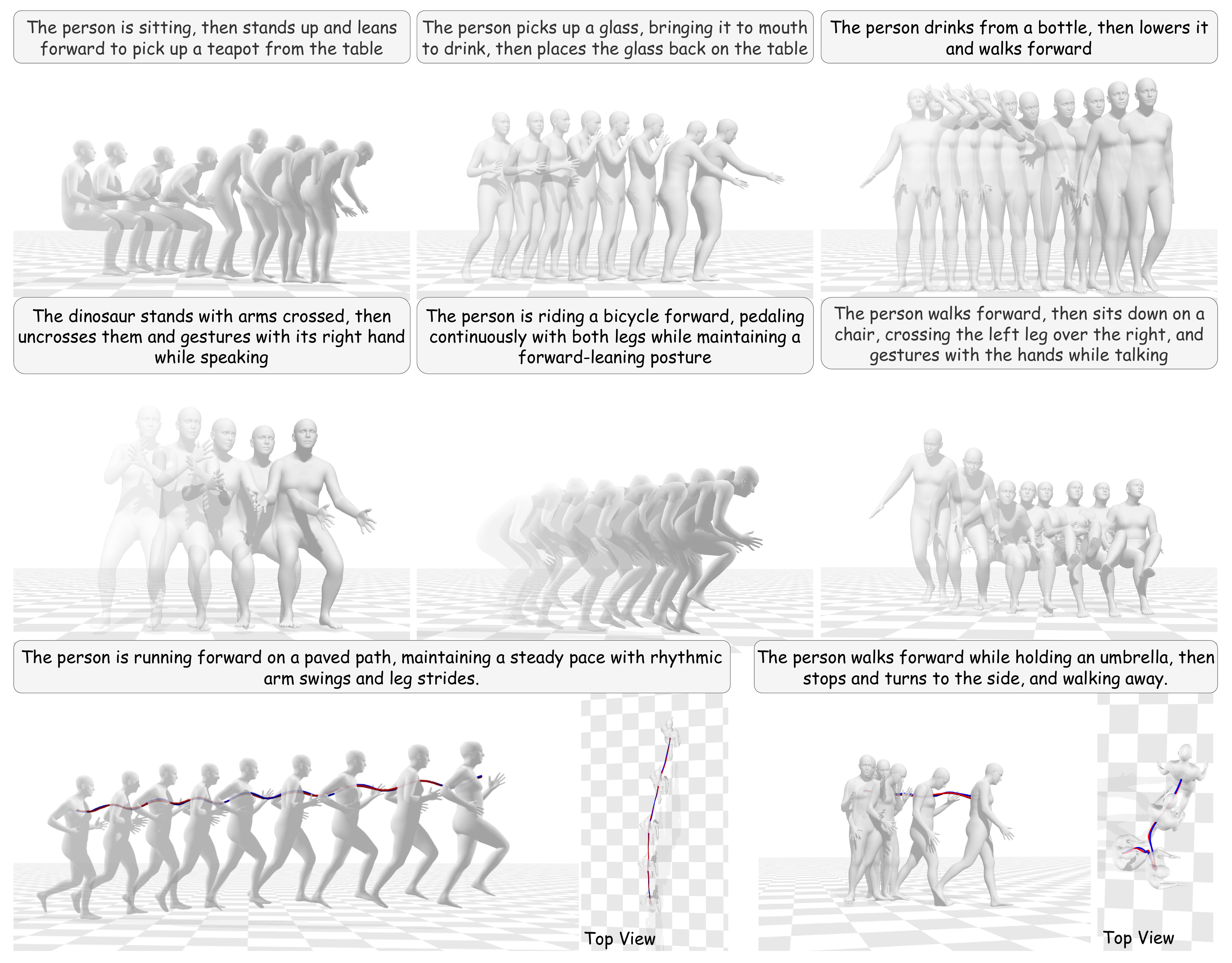}
    \caption{Visualization on text-driven motion generation task.}
    \label{fig:t2m_vis}
\end{figure}

\begin{figure}[t]
    \centering
    \includegraphics[width=\linewidth]{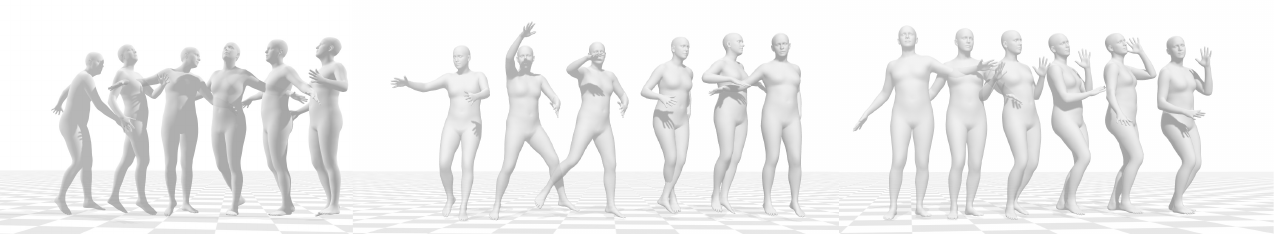}
    \caption{Visualization on music-driven motion generation task.}
    \label{fig:m2d_vis}
\end{figure}

\begin{figure}[t]
    \centering
    \includegraphics[width=\linewidth]{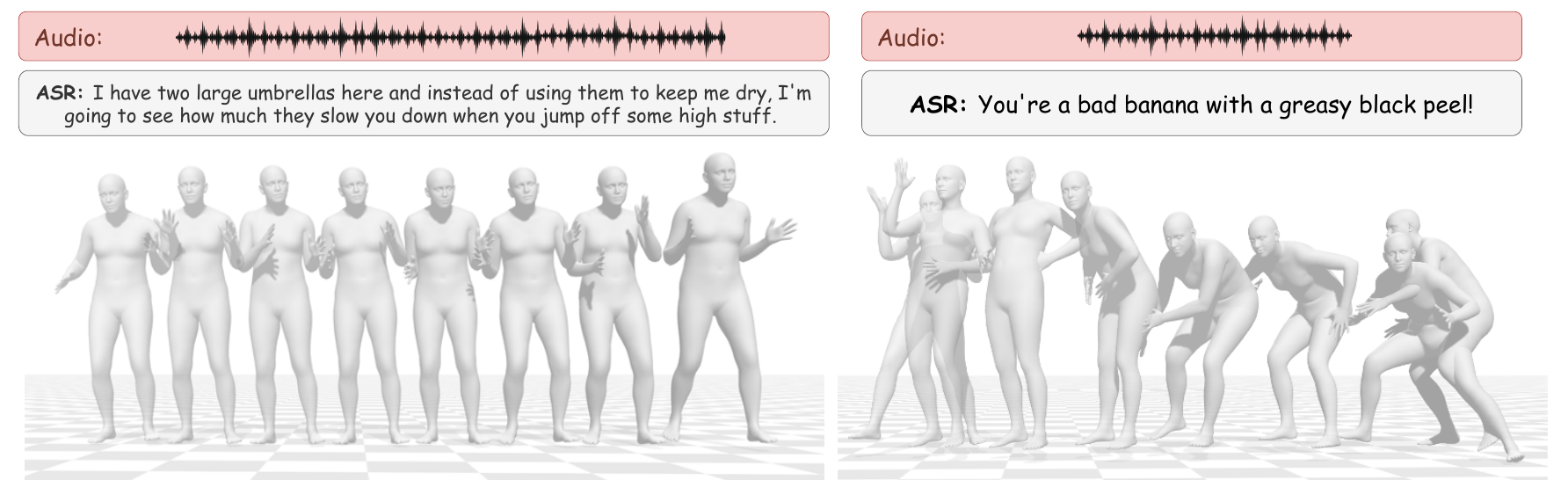}
    \caption{Visualization on speech-driven motion generation task.}
    \label{fig:s2g_vis}
\end{figure}

We present additional examples to visualize motions generated by AnyMo, as shown in Fig. \ref{fig:t2m_vis}, \ref{fig:m2d_vis}, and \ref{fig:s2g_vis}. For trajectory-controlled generation, following TLControl \cite{wan2024tlcontrol}, we adopt a test-time optimization strategy to refine coarse predictions for more precise trajectory control. These visualization results demonstrate that our model produces motion sequences that closely follow diverse input modalities.

\section{Limitation and Future works}
\label{sec:limitation}
Despite the advances demonstrated in this work, several limitations remain and open avenues for future research. First, OmniHuMo does not include finger joint annotations. Compared with body movements, hand regions in internet videos are more frequently affected by occlusion and motion blur, resulting in extremely low data usability for reliable hand reconstruction. Second, audio-aligned data accounts for only a limited portion of the dataset. Future work could improve data diversity and coverage through more targeted data collection strategies.

\clearpage